\documentclass[lettersize,journal]{IEEEtran}
\usepackage{amsmath,amsfonts}
\usepackage{algorithmic}
\usepackage{array}
\usepackage[caption=false,font=normalsize,labelfont=sf,textfont=sf]{subfig}
\usepackage{textcomp}
\usepackage{stfloats}
\usepackage{url}
\usepackage{verbatim}
\usepackage{cite}
\usepackage{pifont}
\usepackage{bbm}
\usepackage{amsmath}
\usepackage{amssymb}
\usepackage{booktabs}
\usepackage{multirow}
\usepackage{xcolor}
\usepackage{url}
\usepackage{caption}
\usepackage{subcaption}

\usepackage{colortbl}
\definecolor{mygray}{rgb}{1,0.95,0.97}

\newcommand{\ie}{\emph{i.e.,}~}

\definecolor{sgreen}{RGB}{30, 150, 30} 
\definecolor{cvprblue}{RGB}{54,125,189}

\usepackage[pagebackref,breaklinks,colorlinks,allcolors=blue]{hyperref}

\usepackage[capitalize]{cleveref}
\crefname{section}{Sec.}{Secs.}
\Crefname{section}{Section}{Sections}
\Crefname{table}{Table}{Tables}
\crefname{table}{Tab.}{Tabs.}

\usepackage{bm}
\usepackage{amsthm,amsmath,amssymb}
\usepackage{mathrsfs}
\usepackage{booktabs}
\usepackage{multirow}
\usepackage{float}
\usepackage{subfig}
\usepackage[table]{xcolor}
\usepackage{multirow}
\usepackage{enumitem}
\usepackage{bbding}
\usepackage{mathrsfs}
\usepackage{color}
\usepackage{colortbl}
\usepackage{rotating}
\usepackage{array}

\usepackage{adjustbox}

\usepackage{tabularx}
\usepackage{etoolbox}
\usepackage{makecell}
\usepackage{pifont}

\usepackage{wrapfig}
\usepackage{cutwin}
\usepackage{alltt}
\usepackage{graphicx}
\usepackage{multirow}
\usepackage{enumitem}

\usepackage{mathrsfs}
\usepackage{booktabs}
\usepackage{multicol}
\usepackage{float}
\usepackage{subfig}
\usepackage{colortbl}
\usepackage{rotating}
\usepackage{array}
\usepackage{orcidlink}

\definecolor{LightCyan}{rgb}{0.88,1,1}

\usepackage[ruled]{algorithm2e}

\SetAlgoNlRelativeSize{0}

\begin{document}

\title{ViC-Bench: Benchmarking Visual-Interleaved Chain-of-Thought Capability in MLLMs with Free-Style Intermediate State Representations}

\author{Xuecheng Wu\orcidlink{0000-0002-6244-0269}, Jiaxing Liu, Danlei Huang\orcidlink{0009-0008-5929-6455}, Yifan Wang, Yunyun Shi\orcidlink{0009-0003-7320-7934}, Kedi Chen, Junxiao Xue\orcidlink{0000-0003-1569-5362}, \\ 
\hspace*{37pt}
Yang Liu\orcidlink{0000-0002-1312-0146}, 
Chunlin Chen\orcidlink{0009-0000-0690-9977},~\IEEEmembership{Fellow, IEEE}, 
Hairong Dong\orcidlink{0000-0002-7255-2950},~\IEEEmembership{Fellow, IEEE}, Dingkang Yang\orcidlink{0000-0003-1829-5671}
\thanks{This research was supported by the Zhejiang Provincial Natural Science Foundation of China under Grant No. LQ23F030009 and the Key R\&D Program of Zhejiang under Grant No. 2024C01036.}
\thanks{Xuecheng Wu, Danlei Huang, and Yunyun Shi are with the School of Computer Science and Technology, Xi'an Jiaotong University, Xi'an, 710049, China. (E-mail: {\small wuxc3@stu.xjtu.edu.cn});}
\thanks{Jiaxing Liu is with Meituan Inc., Shanghai, 200082, China. (E-mail: {\small liujiaxing10@meituan.com});}
\thanks{Yifan Wang is with the Institute of Advanced Technology, University of Science and Technology of China, Hefei, 230031, China (E-mail: {\small wangyfan@mail.ustc.edu.cn});}
\thanks{Kedi Chen is with the School of Computer Science and Technology, East China Normal University and Shanghai Innovation Institute, Shanghai, 200062, China (E-mail: {\small kdchen@stu.ecnu.edu.cn});}
\thanks{Junxiao Xue is with the Research Center for Space Computing System, Zhejiang Lab, Hangzhou, 311100, China (E-mail: {\small xuejx@zhejianglab.cn});}
\thanks{Yang Liu and Hairong Dong are with the College of Electronic and Information Engineering, Tongji University, Shanghai, 201804, China. (E-mails: yang\_liu@ieee.org, hrdong@tongji.edu.cn).}
\thanks{Chunlin Chen is with the School of Robotics and Automation, Nanjing University, Nanjing, 215163, China (E-mail: {\small clchen@nju.edu.cn});}
\thanks{Dingkang Yang is with the College of Intelligent Robotics and Advanced Manufacturing, Fudan University~\&~Fysics AI, Shanghai, 200433, China (E-mail: {\small dkyang20@fudan.edu.cn});}
\thanks{Xuecheng Wu \& Jiaxing Liu deserve equal contributions.}
\thanks{Corresponding authors: Dingkang Yang \& Yang Liu.}
\thanks{Manuscript received XX, 2025; revised XX, 2026.}
}

\markboth{IEEE Transactions on Image Processing}%
{Shell \MakeLowercase{\textit{et al.}}: A Sample Article Using IEEEtran.cls for IEEE Journals}

\maketitle

\begin{abstract}
Visual-Interleaved Chain-of-Thought (VI-CoT) enables Multi-modal Large Language Models (MLLMs) to continually update their understanding and decision space based on step-wise intermediate visual states (IVS), much like a human would, which has demonstrated impressive success in various tasks, thereby leading to emerged advancements in related downstream benchmarks. Despite promising progress, current benchmarks provide models with relatively fixed IVS, rather than free-style IVS, whch might forcibly distort the original thinking trajectories, failing to evaluate their intrinsic reasoning capabilities. More importantly, existing benchmarks neglect to systematically explore the impact factors that IVS would impart to the untamed reasoning performance. To tackle above gaps, we introduce a specialized benchmark termed ViC-Bench, consisting of four representive tasks, \ie maze navigation, jigsaw puzzle, embodied long-horizon planning, as well as complex counting, where each task has dedicated free-style IVS generation pipeline supporting adaptive function calls. To systematically examine VI-CoT capability, we propose a thorough evaluation suite incorporating a progressive three-stage strategy with targeted new metrics. Besides, we establish Incremental Prompting Information Injection strategy to ablatively explore the prompting factors for VI-CoT. We extensively conduct evaluations for 18 advanced MLLMs, revealing key insights into their VI-CoT capability. The introduced ViC-Bench has been made publicly available at~\href{https://huggingface.co/datasets/meituan/ViC-Bench}{Huggingface}.
\end{abstract}


\section{Introduction}
\label{sec:intro}

Multi-modal AI field is currently evolving from LLMs~\cite{LLM_review-1,LLM_review-2,Llama3_touvron2023llama} to MLLMs~\cite{MLLM_review-1,TIP-VLMs-2,TIP-VLMs}, which integrate various modalities into the backend language decoders~\cite{Multimodal-chain-of-thought}. Achieving human-level multi-modal intelligence requires transcending basic perceptual capabilities to attain sophisticated reasoning. Drawing inspirations from the remarkable success of Chain-of-Thought (CoT) in LLMs~\cite{Towards-reasoning-era,Mind-with-eyes,Deepseek-r1}, the integration of CoT into visual-language contexts has catalyzed transformative progress, giving rise to visual CoT~\cite{li2025imagine-CoT,Multimodal-chain-of-thought}.

The initial visual CoT involves vision signals only as input, whereas the entire rationales are composed of language, in which various methods and related benchmarks make rapid advancements~\cite{Multi-Evaluation,VERIFY,xu2024LLaVA-CoT}. However, this paradigm overlooks the explicit visual representation updates and continuous understanding of visual feedbacks, misaligning with the human cognitive process of using visual thoughts for concrete reasoning and textual thoughts for abstract reasoning. To this end, Visual-Interleaved Chain-of-Thought (VI-CoT), which incorporates step-wise intermediate visual states (IVS) based on visual inputs, has made rapid progress. According to the source of IVS, current VI-CoT methods are primarily divided into two types. The first type involves autonomously generating IVS based on its internalized understanding~\cite{li2025imagine-CoT,RISEBench}. However, this approach currently struggles due to the limited generative capabilities of MLLMs~\cite{Multi-step,RISEBench}. The second type involves providing IVS through external knowledge retrieval, utilizing expert tools or human in the agent-form, which shows impressive results in various tasks~\cite{CMMCoT,Comt_AAAI,Embodied-agent-interface}. Meanwhile, to evaluate the developments of recent VI-CoT methods, various benchmarks have emerged~\cite{Comt_AAAI,magebench}. Despite promising advancements, few of them provides free-style IVS representations to MLLMs, as illustrated in Tab.~\ref{tab-intro-compare} below. CoMT~\cite{Comt_AAAI} primarily provides fixed IVS, which could forcibly distort the original planning trajectories of models. While MageBench~\cite{magebench} offers the dynamic IVS but imposes the attribute constraints of action-observation memory. More importantly, existing benchmarks~\cite{LEGO-Puzzles,MME-CoT,Comt_AAAI,VERIFY} neglect to systematically assess the impact factors that IVS would impart to the untamed reasoning performance in MLLMs. (\textit{i.e.}, Positive, Negative, or Null). As a result, a natural question arises: \textit{Could MLLMs leverage VI-CoT, which closely aligns with human cognitive behavior, to inherently achieve better reasoning performance?}

\begin{figure}[t!]
\centering
\includegraphics[width=\linewidth]{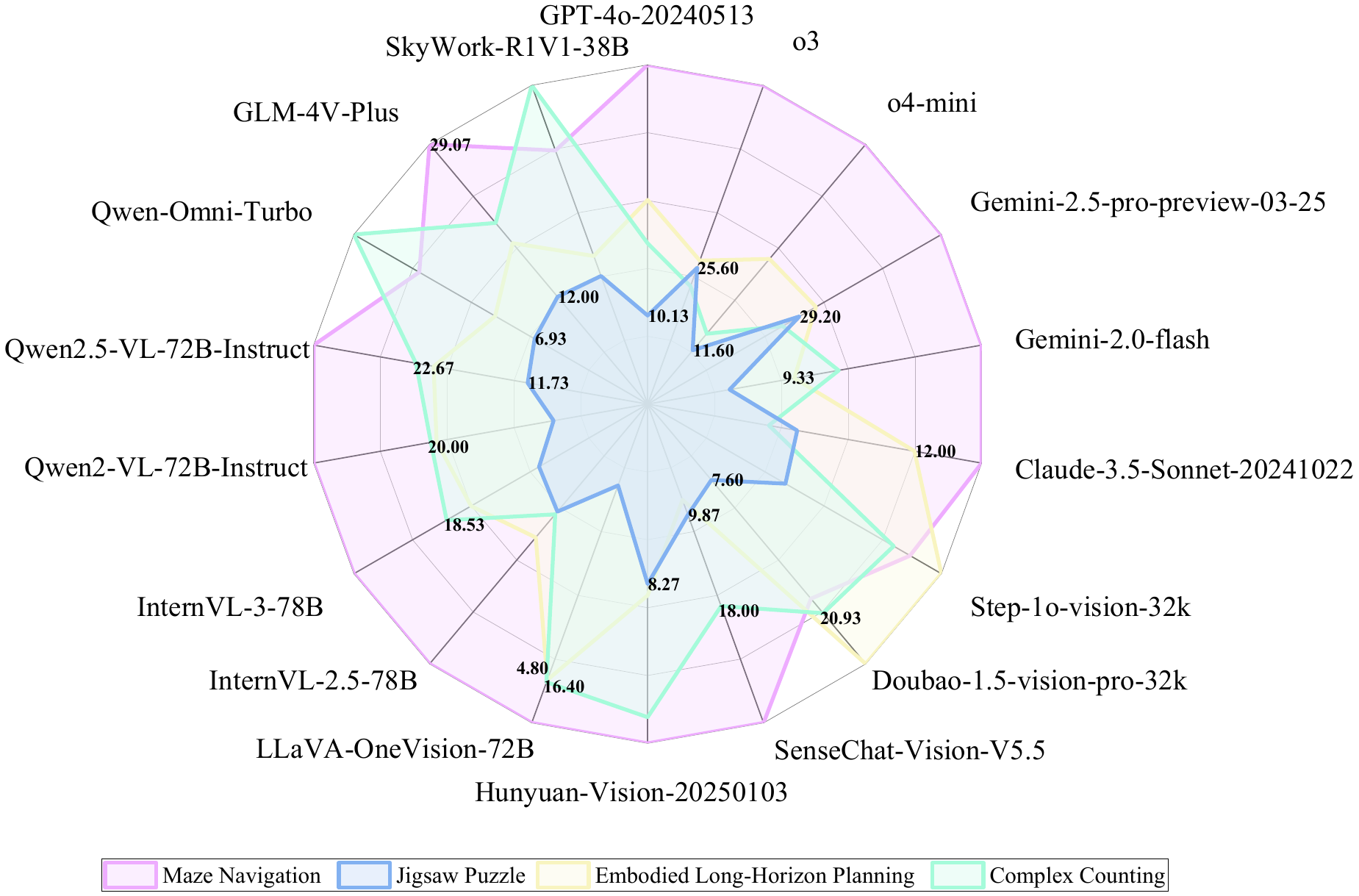}
\vspace{-0.3em}
\caption{Performance overview of advanced MLLMs on four tasks in terms of \textit{average} \textit{ACC} across three evaluation stages.}
\vspace{-1.45em}
\label{radar-fig}
\end{figure}

To tackle above gaps, we introduce the specialized ViC-Bench for evaluating VI-CoT, selecting four representative tasks (\textit{i.e.}, maze navigation, jigsaw puzzle, embodied long-horizon planning, and complex counting), which require models to dynamically interact with visual contexts and continuously update their understanding and decision-making based on step-wise IVS. We first propose dedicated data construction pipelines, resulting in 250 unique images for each task. We then introduce free-style IVS generation workflows supporting function calls to sufficiently support the investigation on VI-CoT performance. Based on the constructed data, we propose a novel evaluation suite incorporating progressive three-stage evaluation strategy with targeted new metrics. Specifically, Stage 1 involves multiple-choice QA, while Stage 2 focuses on open-ended QA, both utilizing visual signals solely at the input to establish solid foundations for Stage 3. Building on this, Stage 3 features open-ended QA with free-form IVS, enriching from the hierarchical references for VI-CoT evaluation through our progressive design. Subsequently, we define new Recall and ThinkGain metrics by black-boxing the retrospection process, along with a Legality metric to tackle the clear rule boundaries, thereby establishing a comprehensive suite for each stage. To ablatively explore the prompting factors affecting VI-CoT capability, we further design the Incremental Prompting Information Injection (IPII) strategy across three stages, utilizing varying global-aware prompting levels. We totally examine 18 advanced MLLMs on introduced ViC-Bench, providing extensive quantitative and qualitative results. The performance overview is illustrated as Fig.~\ref{radar-fig} above. We uncover significant performance gaps between open-source and proprietary MLLMs in both quantitative analyses and prompting studies. Moreover, most MLLMs show substantial disparities compared to human-level proficiency.

In addition to providing insights into VI-CoT capabilities of MLLMs, ViC-Bench can also establish foundations for the progress of unified MLLMs~\cite{Emu3d5,bagel}, multi-modal agents~\cite{multi-modal-agent-1,multi-modal-agent-2}, embodied AI~\cite{Embodied-agent-interface,system2-review1}, and autonomous driving~\cite{autonomous-driving}. In summary, the main contributions of this paper are three-fold:
\begin{itemize}[leftmargin=*]
\item[$\bullet$]We curate ViC-Bench, including four representative tasks, each with dedicated construction and free-style IVS generation pipelines. Moreover, we engage human-machine collaborations in both construction and evaluation to establish a high-quality benchmark for VI-CoT reasoning.
\item[$\bullet$]We propose a thorough evaluation suite that includes a progressive three-stage evaluation strategy with newly targeted metrics to meticulously examine the inherent VI-CoT performance using free-style IVS. We further introduce the Incremental Prompting Information Injection (IPII) strategy to ablatively explore the prompting factors for VI-CoT.
\item[$\bullet$]Extensive experiments and analyses are performed on 18 proprietary and open-source MLLMs. We summarize several key observations and insights, hoping to inspire advancements of future research.
\end{itemize}

\begin{table*}[t!]
\centering
\caption{The statistics comparisons of ViC-Bench and related representative benchmarks. \#No.: Unique sample number. Free-Style: Free-style IVS supporting function calls.}
\vspace{-0.4em}
\setlength{\arrayrulewidth}{0.40pt}
\renewcommand{\arraystretch}{1.00}
\resizebox{0.99\textwidth}{!}{%
\begin{tabular}{lcccccccc}
\toprule
Benchmark & Venue  & Source & \#No.  & Task & Multi-Step & IVS & Free-Style \\
\midrule

M$^3$CoT~\cite{m3cot} (Test Part)   & ACL'24   & Web & 2,358   & 3   & \checkmark & $\times$  & $\times$    \\
LEGO-Puzzles~\cite{LEGO-Puzzles} & arXiv'25  & Synthesized   & 1,100  & 3   & \checkmark & $\times$  & $\times$     \\
MME-CoT~\cite{MME-CoT}    &  ICML'25   & Web   & 808   & 6   & \checkmark & $\times$  & $\times$    \\
CoMT~\cite{Comt_AAAI}    & AAAI'25  & Web   & 3,853   & 4 & \checkmark & \checkmark   & $\times$    \\
MageBench~\cite{magebench}    & arXiv'24  & Synthesized \& Web   & 483   & 3   & \checkmark & \checkmark   & \checkmark    \\
VERIFY~\cite{VERIFY} & arXiv'25   & Web   & 600  & 1   & $\times$ & $\times$ & $\times$   \\

\rowcolor{gray!20}
\textbf{ViC-Bench (Ours)} & \textbf{--} & Synthesized \& Web   & 2,751   & 4   & \checkmark & \checkmark  & \checkmark    \\

\hline
\end{tabular}
}
\label{tab-intro-compare}
\vspace{-1.1em}
\end{table*}

\section{Related Work}
\label{sec:Related Work}

\subsection{Multi-modal Large Language Models}
The integration of multi-modal information with LLMs~\cite{LLM_review-1,LLM_review-2} leads to the emergence of MLLMs~\cite{MLLM_review-1,TIP-VLMs-2,TIP-VLMs}, exhibiting impressive performance in various multi-modal understanding and generation tasks. MLLMs can be broadly categorized into two types: pipeline-based and native paradigms. The pipeline-based MLLMs can be generally classified into three types based on the multi-modal integration strategies: (1) Feature mapping with MLPs, such as PaLM-E~\cite{30_driess2023palm}, LLaVA~\cite{29_liu2024visual}, and CogVLM~\cite{32_wang2023cogvlm}; (2) Query-based cross-attention components (\textit{e.g.}, InstructBLIP~\cite{4_dai2023instructblip}, Mini-GPT4~\cite{26_zhu2023minigpt}, and Qwen-VL series~\cite{28_bai2023qwen,Qwen2-vl,Qwen2d5-vl}); (3) Cross-attention layers within LLMs, such as Flamingo~\cite{35_alayrac2022flamingo} and IDEFICS~\cite{37_laurenccon2024obelics}. Meanwhile, MLLMs integrated with generation capabilities through coupled components have also been largely promoted~\cite{44_peng2023kosmos,Emu,Minigpt-5}. As for native MLLMs, they primarily achieve unified understanding and generation through auto-regressive manners with elaborate tokenizers~\cite{Liquid,LWM2024,Chameleon}. Most recently, the release of OpenAI o3/o4~\cite{openai2025o4mini} and DeepSeek-R1~\cite{Deepseek-r1} sparks a wave of interests in reasoning enhancements, highlighting the effectiveness of CoT~\cite{system2-review1,zeng2024mr-Mr-ben}. Inspired by this, researchers have sought to advance the reasoning capabilities of MLLMs by employing visual CoT mechanisms.~\cite{MM-Eureka,Vlm-r1,liu2024enhancing-CoT-ZJU}.

\subsection{Visual-Interleaved Chain-of-Thought}
VI-CoT involves the engagement of step-wise IVS through the entire reasoning process, achieving impressive performance across various downstream scenarios. However, due to the limited visual generative capabilities, MLLMs struggle to generate the native IVS, which are essential for the in-context knowledge retrieval. As a result, current methods primarily rely on external knowledge retrieval to develop IVS, such as expert tools or human in the agent-form~\cite{Multimodal-chain-of-thought,Mind-with-eyes}. CMMCoT~\cite{CMMCoT} utilizes the visual region tokens as supervisory signals to perform interleaved reasoning. Zhang et al.~\cite{Embodied-reasoner} extend o1-style reasoning to interactive embodied search. Gao et al.~\cite{gao2024interleaved-CoT} propose the attention-driven selection method to realize interleaved CoT. VoT~\cite{Video-of-thought} breaks down complex task into sub-problems, and address them from low to high employing scene graphs. MVoT~\cite{li2025imagine-CoT} enables visual thinking by generating visual visualizations of reasoning trajectories. Hu et al.~\cite{Visual-sketchpad} provide MLLMs with sketchpad and expert tools to conduct interleaved CoT. Meanwhile, related benchmarks have emerged to extensively evaluate various VI-CoT methods. CoMT~\cite{Comt_AAAI} constructs four types of visual operations, requiring multi-modal reasoning outputs. Zhang et al.~\cite{magebench} propose MageBench for evaluating the MLLMs’s capabilities of being an agent. \cite{Embodied-reasoner} totally cultivates 809 test cases across 12 scenarios for hierarchical embodied long-horizon tasks. Despite great advancements, few of these benchmarks provide the free-form IVS representations and systematically evaluate the influence that IVS can make on the untamed reasoning performance. To bridge these gaps, we carefully construct four representative VI-CoT tasks with the free-form IVS representations and further propose a comprehensive evaluation suite incorporating three progressive stages.

\section{Benchmark Construction}
\label{sec:bench-construction}

\subsection{Overview}
We introduce ViC-Bench, comprising four representative VI-CoT tasks, \ie Maze Navigation~(Sec.~\ref{Maze}), Jigsaw Puzzle~(Sec.~\ref{Puzzle}), Embodied Long-Horizon Planning~(Sec.~\ref{embodied}), and Complex Counting~(Sec.~\ref{counting}). The overall construction workflow can be mainly divided into Raw Data \& Pre-processing, Three-Stage Construction, IVS Generation, and Human Recheck, as illustrated in Fig.~\ref{fig:pipelines}.

\subsection{Maze Navigation}
\label{Maze}

\noindent \textbf{Raw Data \& Pre-processing.}~As shown in Fig.~\ref{fig:pipelines}~(a), we utilize Maze~\cite{maze-dataset} library coupled with DFS method to render 4~$\times$~4 mazes in multiple batches. After generation, we first screen out those with navigation lengths ranging from 5-8, then aggregate mazes with the same starting point. Finally, mazes with duplicate shortest path after global-aware aggregation are removed to ensure uniqueness.

\noindent \textbf{Stage 1.}~For the processed mazes, we randomly select an original maze $\mathbf{M_1}$ in the non-overlapping paradigm, then select three other distinct mazes $\mathbf{M_2}$, $\mathbf{M_3}$, and $\mathbf{M_4}$. Subsequently, we draw the endpoints of $\mathbf{M_2}$-$\mathbf{M_4}$ onto $\mathbf{M_1}$ to establish three incorrect options serving as visual distractors, facilitating MLLMs to select the correct endpoint based on the starting point and the given navigation path.

\noindent \textbf{Stages 2 \& 3.}~Based on $\mathbf{M_1}$ of Stage 1, we mark the starting and corresponding endpoints with $\mathbf{S}$ and $\mathbf{E}$, and require models to respond with the correct path from $\mathbf{S}$ to $\mathbf{E}$ under the clear rules. Stage 3 further builds upon Stage 2, with its main feature being the application of free-style IVS in response to the step-wise instructions of models under the agent-form, promoting to investigate the untamed VI-CoT boundaries.

\noindent \textbf{IVS Generation.}~Based on simulated functions, we perform multi-step simulations on the input maze, employing a blue pentagram to mark the agent position. We set the maximum attempts to 30.

\noindent \textbf{Human Recheck.}~Throughout three-stage data construction, we employ human experts to perform one-by-one recheck on the 250 mazes to sufficiently ensure data feasibility. Besides, we conduct manual quality inspections on the meta outcomes of Stage 3 to faciliate the stability of our evaluations.

\subsection{Jigsaw Puzzle}
\label{Puzzle}

\noindent \textbf{Raw Data \& Pre-processing.}~To eliminate the risk of data leakage from the familiar datasets~\cite{56_deng2009imagenet,57_lin2014microsoft}, we construct an image source pool using elaborate prompts with DALLE-3~\cite{DALLE-3}, FLUX.1-schnell~\cite{flux2024}, Kwai-Kolors~\cite{kolors}, Stable-Diffusion-3 Medium~\cite{stablediffusion}, WanX~\cite{WanX}, and Midjourney-V6.1~\cite{Midjourney}, which can sufficiently ensure the image diversity and uniqueness. Following~\cite{t2i-eva-1,t2i-eva-2}, we employ the manual white-box approach to filter the generated images, primarily considering three discriminative metrics, \textit{i.e.}, T2I consistency, reasonableness, and, which can be formulated as:
\begin{equation}
S_{overall} = 0.6 \times S_{cy} + 0.2 \times S_{rs} + 0.2 \times S_{rm},
\label{puzzle-metric}
\end{equation}
where $S_{cy}$, $S_{rs}$, and $S_{rm}$ denote the scores of consistency, reasonableness, as well as realism, respectively. Moreover, $S \in [1, 5] \cap \mathbb{Z}$. Based on the descending order of overall scores, we finally select 250 unique images.

\begin{figure*}[t!]
\centering
\includegraphics[width=\linewidth]{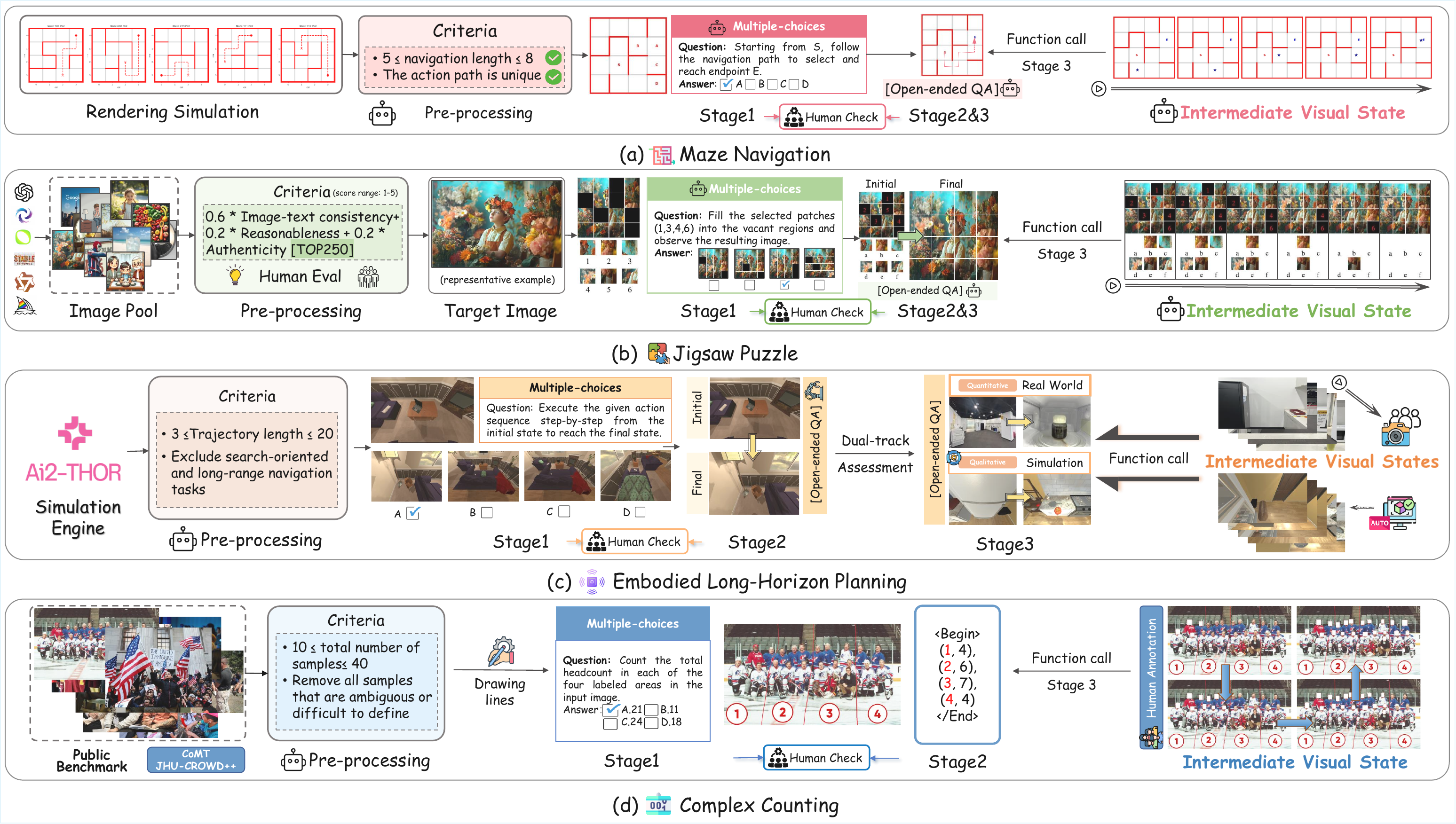}
\vspace{-1.5em}
\caption{The illustrations of the overall construction pipelines for four representative VI-CoT tasks in proposed ViC-Bench.}
\vspace{-1.4em}
\label{fig:pipelines}
\end{figure*}

\noindent \textbf{Stage 1.}~We first adjust the selected images to 224 $\times$ 224 resolutions and divide them into 4 $\times$ 4 patches. We then targetedly extract six patches using proposed weighted dispersed sampling strategy (as described in Sec.~\textcolor{blue}{IV} of supplementary material). Afterwards, the selected patches are removed from the original image to generate masked image \textit{$\mathbf{I}_m$}. Subsequently, the chosen patches are randomly numbered and concatenated with \textit{$\mathbf{I}_m$} along the vertical direction to output the overall input image, as illustrated in Fig.~\ref{fig:pipelines}~(b). As for options, four patches are first randomly selected and correctly positioned to establish the correct option. Three incorrect options are then generated based on the puzzle states with either (3 \checkmark \& 1 $\times$) or (2 \checkmark \& 2 $\times$) patch placements, leading to the plausible yet incorrect paradigm, which challenge the discriminative capabilities of MLLMs in region-aware semantic understanding and cross-level spatial reasoning.

\noindent \textbf{Stages 2 \& 3.}~The overall input images of the last two evaluation stages almost keep the same with Stage 1, but additionally label the six vacant regions in \textit{$\mathbf{I}_m$}. Moreover, the arrangement of six patches in the overall input image underneath remains consistent with Stage 1. The task instructions here require models to directly output the correct mapping between patches and vacant regions.

\noindent \textbf{IVS Generation.}~Based on the simulated functions and planning trajectories of MLLMs, we conduct multi-step agent-form evaluations with free-style IVS. Besides, we establish the rule boundaries such that if a patch or vacant region is repeatedly utilized, causing conflicts, we will deem the current action invalid and provide appropriate guidance. To keep consistent with maze navigation, we set the maximum attempts to 30.

\noindent \textbf{Human Recheck.}~Following maze navigation, we also employ human experts to recheck the generated puzzles across all stages and perform thorough quality inspections on VI-CoT evaluations in Stage 3.

\subsection{Embodied Long-Horizon Planning}
\label{embodied}

\noindent \textbf{Raw Data \& Pre-processing.}~We construct the dataset for this task based on the AI2-THOR simulation environment~\cite{ai2thor}. To ensure the task difficulty is appropriately calibrated for long-horizon planning, we filter action sequences based on trajectory length, retaining only those with a step count in the range of $3$ to $20$. Furthermore, to mitigate the issue of unreachable target locations caused by navigation errors within the simulator, we rigorously exclude search-oriented tasks. Specifically, tasks involving ``PickUp'', ``Put'', and other long-range navigation objectives are systematically removed to guarantee the validity and plausibility of the ground truth paths. In the end, we manually curate a diverse and high-quality subset of 250 samples for final evaluation.

\noindent \textbf{Stage 1.}~We directly regard the final visual state achieved after completing the overall action sequence as the correct option. Three incorrect options are generated using ambiguous IVS, which are extracted from the original execution trace and manually refined to increase perceptual and semantic difficulty, as displayed in Fig.~\ref{fig:pipelines}~(c). All options are further verified by human experts to ensure plausibility. This setup thoroughly evaluates the long-horizon planning capabilities of models and provides strong baselines for subsequent evaluations.

\noindent \textbf{Stage 2.}~For each sample, we take the initial observation of the action sequence from Stage 1 as the starting state $\mathbf{O}_s$, and the fully achieved final state as the ending state $\mathbf{O}_e$. In this stage, MLLMs are required to open-endedly generate an action sequence that successfully transitions from $\mathbf{O}_s$ to $\mathbf{O}_e$.

\noindent \textbf{Stage 3 \& IVS Generation.}~In this stage, we leverage the rendering engine~\cite{ai2thor} to enable dynamic evaluation. Specifically, each action predicted by models is rendered and executed within the environment. An action is deemed legal only if it is successfully executed, thereby strictly penalizing hallucinated or physically infeasible operations. Complementing this quantitative evaluation, we further introduce a human-in-the-loop paradigm for qualitative assessment in real-world settings. Following~\cite{magebench}, we adopt a collaborative framework in which the MLLM serves as the \textit{planner} and the human acts as the \textit{executor}. Within this setup, human experts capture intermediate visual states (IVS) based on the model's responses and upload them to the cloud via a dedicated mobile APP. These IVS are then utilized in conjunction with function calling to enable step-wise, human-machine collaborative evaluation.

\subsection{Complex Counting}
\label{counting}

\noindent \textbf{Raw Data \& Pre-processing.}~We build this subset based on the samples from JHU-CROWD++~\cite{Jhu-crowd++} and CoMT~\cite{Comt_AAAI}. According to the data annotations in \cite{Comt_AAAI}, we first filter out samples with headcounts ranging from 10 to 40, then remove those that are ambiguous or difficult to delineate. As displayed in Fig.~\ref{fig:pipelines}~(d), we then manually draw irregular lines to separate the input image into four regions, leading to the completed input image \textit{$\mathbf{H}$}. In the end, we construct 250 diverse samples.

\noindent \textbf{Stage 1.}~We manually count the visible heads and regard the correct headcounts as the ground truth option, tackling limitations in previous work~\cite{Comt_AAAI} that solely relies on DNNs for annotations. We then employ human experts to establish three incorrect options, leading to higher selection difficulties.

\noindent \textbf{Stages 2 \& 3.}~The input image remains consistent with Stage 1, but the QA format changes from multiple-choice to open-ended, where we require models to directly respond with the headcounts for each region. The output format is defined as \textless Begin\textgreater~($\mathbf{1}$, $\mathbf{h_1}$), ($\mathbf{2}$, $\mathbf{h_2}$), ($\mathbf{3}$, $\mathbf{h_3}$), ($\mathbf{4}$, $\mathbf{h_4}$)~\textless/End\textgreater, where $\mathbf{h_i}$ ($\mathbf{i} \in \{1, 2, 3, 4\}$) denotes the headcount in each region.

\noindent \textbf{IVS Generation.}~Considering that complex counting lacks the explicit action dynamics present in tasks like maze navigation or jigsaw puzzle, and empirical results further indicate that overly fine-grained masks can induce overthinking and object hallucinations, we thus utilize the coarse-grained region-aware masks instead of fine-grained per-head masks for IVS generation. Specifically, for each region, we manually draw a number of bounding boxes according to the model's predicted head count, regardless of whether this matches the ground-truth number. This process yields sequentially stacked masks that can fully cover, under-cover, or over-cover the actual heads, thereby reflecting the model's counting behavior.

\begin{figure*}[t!]
\centering
\includegraphics[width=\textwidth]{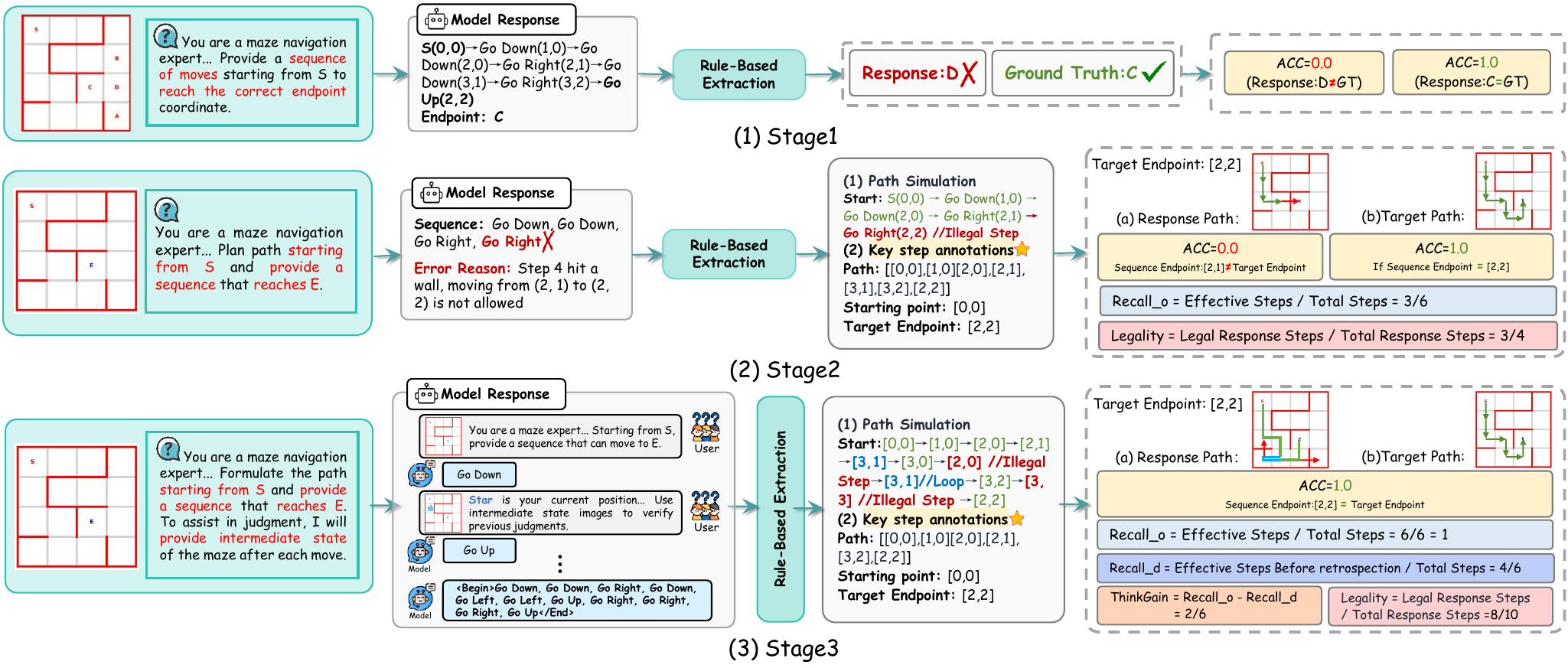}
\vspace{-1.5em}
\caption{The overall illustration of our progressive three-stage evaluation strategy taking maze navigation as the example.}
\vspace{-1.2em}
\label{fig:pipelines-eva}
\end{figure*}

\section{Evaluation Suite}
\label{sec:bench-Suite}

\subsection{Progressive Three-Stage Evaluation Strategy}
\label{3stages}

As illustrated in Fig.~\ref{fig:pipelines-eva}, we propose a holistic progressive three-stage evaluation strategy, rather than simply focusing on the final answer, to examine the untamed understanding of VI-CoT capability in MLLMs. Remarkably, we take the maze navigation task as an example to demonstrate the key steps in the entire workflow. Stage 1 only focuses on the final results, employing the common multiple-choices QA~\cite{Scienceqa,ding2024mmvqa,76_masry2022chartqa} to preliminarily determine the visual CoT performance. Based on POMDP~\cite{magebench}, Stage 1 can be formulated as:
\begin{equation}
\pi_\theta(p_{sys}, (Q, \mathbf{v_0}, C), p_{task}, p_{cot}, p_{io})\rightarrow (\mathbf{r_1}, \mathbf{r_2}, \mathbf{r_3}, ..., \mathbf{r_T}, \mathbf{C_S}),
\label{stage1}
\end{equation}
where $p_{sys}$ and $p_{task}$ denote system and task prompts. $p_{cot}$ and $p_{io}$ refer to CoT and IO prompts, designating the inner thought flow and output format. $(Q, \mathbf{v_0}, C)$ represents question and initial observation with options. $\mathbf{r_i}$ ($\mathbf{i} \in \{1, ..., \mathbf{T}\}$) are the reasoning rationales. $\mathbf{C_S}$ is the selection. In Stage 2, we convert the multiple-choices QA into the more challenging open-ended format, \textit{i.e.},
\begin{equation}
\pi_\theta(p_{sys}, (Q, \mathbf{v_0}), p_{task}, p_{cot}, p_{io})\rightarrow (\mathbf{r_1}, \mathbf{r_2}, \mathbf{r_3}, ..., \mathbf{r_T}, \mathbf{A_F}),
\label{stage2}
\end{equation}
where $\mathbf{A_F}$ is the formatted answer. Doing so allows us to observe the reasoning performance in an open-ended manner, which leads to direct subjective Answer-Only evaluations, thereby serving as an important indicator for Stage 3. Finally, Stage 3 pays attention to the legal free-style evaluations using function calls in the agent-form to explore in-depth thinking gains brought by the IVS representations, addressing the shortcomings in previous works where fixed IVS might forcefully influence the inherent planning in MLLMs by constraining the judgment path. Overall, this process can be represented as:
\begin{equation}
F(p_{sys}, (Q, \mathbf{v_0}), p_{task}, p_{cot}, p_{io}; \pi_\theta) \rightarrow \mathbf{R},
\end{equation}

\begin{equation}
\begin{split}
\pi_\theta \big( & p_{sys}, (Q, \mathbf{v_0}), ((\mathbf{a_{1}}, \mathbf{v_{1}}), \dots, (\mathbf{a_{t}}, \mathbf{v_{t}})), \\
& p_{task}, p_{cot}, p_{io} \big) \rightarrow (\mathbf{a_{t+1}}, \mathbf{v_{t+1}}),
\end{split}
\label{stage3}
\end{equation}
where $\mathbf{R}$ denotes the final answer and $(\mathbf{a_{t}}, \mathbf{v_{t}})$ refers to action and IVS feedback at the $t^{th}$ step. Our free-style exploration in the agent-form further stimulates the influence of IVS, thereby more comprehensively investigating the untamed VI-CoT capability inherented in advanced MLLMs.

\subsection{Evaluation Metrics}

As shown in Fig.~\ref{fig:pipelines-eva}, Stage 1 employs Accuracy (ACC) metric, Stage 2 includes ACC, Recall\_o, and Legality metrics, Stage 3 further enhances Stage 2 with the newly introduced \textit{ThinkGain} metric.

\noindent \textbf{Stage 1.}~We mainly focus on whether the final choice is correct or not, consistently utilizing ACC as the sole metric across four tasks. As displayed in Fig.~\ref{fig:pipelines-eva} (a), we first perform rule-based extraction to obtain the selected options, then directly compare them with Ground Truth to compute the ACC metric, \textit{i.e.},
\begin{equation}
ACC = \frac{TP + TN}{TP + TN + FP + FN},
\label{eq:accuracy}
\end{equation}
where $TP$, $TN$, $FP$, and $FN$ denote the number of true positives, true negatives, false positives, and false negatives. In this way, we can tentatively observe reasoning performance, offering extensive references.

\noindent \textbf{Stage 2.}~ACC only focuses on the correctness of final answer but neglects how many informative steps the model has taken to reach the correct answer. However, examining the rigorousness of models towards Ground Truth is very crucial, as it can delineate the response progress and provide insights into the effectiveness of open-ended visual CoT in MLLMs. To tackle above gaps, we introduce task-specific Recall\_overall metric (\textit{i.e.}, $\mathbf{R\_o}$), only considering the final output. $\mathbf{R\_o}$ measures the informative steps the model has achieved to emphatically characterize its reasoning ability, \textit{i.e.},
\begin{align}
k_0 &= \underset{k}{\arg \max } \frac{\left|\mathcal{S}_{\text {reached}}^k\right|}{\left|\mathcal{S}^k\right|}, \\
\mathbf{R\_o} &= \frac{\left|\mathcal{S}_{\text {reached}}^{k_0}\right|}{\left|\mathcal{S}^{k_0}\right|},
\end{align}
where $\mathcal{S}$ denotes the key step-wise components for tasks and $\mathcal{S}^k$ refers to the $k^{\text{th}}$ method of a question.

In maze navigation, we extract the final predictions by the rule-based method, then determine the legal endpoint $\mathbf{P_f}$ reached by the predicted path using our rendering simulations. We then compare $\mathbf{P_f}$ with Ground Truth since each constructed maze has a unique shortest path. If $\mathbf{P_f}$ exists in the coordinate set $\mathbf{C_{set}}$ of Ground Truth, we thus determine $\mathcal{S}_{\text {reached}}^k$ based on the distance from $\mathbf{P_f}$ to the starting point $\mathbf{S}$. If $\mathbf{P_f}$ is not in $\mathbf{C_{set}}$, we directly take $\mathbf{R\_o}$ as 0. In jigsaw puzzle, we follow maze navigation to extract final predictions. Considering the uniqueness of puzzle correspondence sequence, we directly perform global-aware strict matching under the component-wise paradigm between predictions and Ground Truth. We calculate $\mathbf{R\_o}$ through taking the number of correctly filled patches as $\mathcal{S}_{\text {reached}}^k$ and the counts of extracted patches as $\mathcal{S}^k$. Considering the atomic action of embodied long-horizon planning task have dependency constraints, we verify the correctness of atomic action utilizing semantic matching~\cite{paddlenlp} rather than strict matching, keeping consistent with~\cite{Embodied-reasoner}. Notably, non-critical steps (\ie observe and move forward) are ignored during matching. A pair-wise match is regarded as successful only when the semantic similarity of atomic action pair exceeds 0.95. For ACC, the sample is taken as correct if all the atomic actions are validly matched. For $\mathbf{R\_o}$, we take the accumulated effective length of verified actions and the total predicted steps as $\mathcal{S}_{\text {reached}}^k$ and $\mathcal{S}^k$, respectively. As for complex counting, we follow jigsaw puzzle to perform global-aware pair-wise matching, except that we further implement an explicit fault-tolerance mechanism due to the high difficulty of this task for current MLLMs, with a tolerance threshold set to 1.

The compliance with explicit rules in maze navigation, jigsaw puzzle, and embodied long-horizon planning tasks poses challenges to the instruction-following and visual perception capabilities of MLLMs, which correlatively promotes VI-CoT evaluation, leading to our \textit{Legality} metric. In maze navigation, we consider two types of illegal behaviors, namely going out of bounds and hitting walls. For jigsaw puzzles, illegal behaviors include repeated patch placements and repeated filling of vacant regions. Regarding embodied long-horizon planning, legal steps necessitate valid actions performed on interactable objects. Specifically, we segment the predictions followed by conducting step-wise simulation rendering to determine the legal steps, \textit{i.e.},
\begin{equation}
\text{Legality} = \frac{S_L}{S_O},
\label{eq:legality}
\end{equation}
where $S_L$ and $S_O$ refers to the number of legal steps and overall partition steps, respectively.

\noindent \textbf{Stage 3.}~Building upon Stages 1 \& 2, we further introduce a new metric denoted ThinkGain to examine the influence of free-style IVS on VI-CoT performance. Drawing inspirations from the reward system of GRPO in \cite{Deepseek-r1}, we black-box the retrospection process and focus only on the decision states ($\mathbf{D_d}$ \& $\mathbf{D_o}$) both before the retrospection commences and after it concludes, avoiding the negative impact of numerous ongoing factors on VI-CoT evaluation. We then employ the Recall metric defined in Stage 2 to assess $\mathbf{D_d}$ and $\mathbf{D_o}$. Overall, the ThinkGain metric can be represented as:
\begin{equation}
\text{ThinkGain} = \mathbf{R\_o} - \mathbf{R\_d},
\label{eq:thinkgain}
\end{equation}
where $\mathbf{R\_d}$ denotes the Recall metric calculated with $\mathbf{D_d}$. Besides, the definition of $\mathbf{D_d}$ varies due to inconsistence in task representations. In maze navigation, we regard $\mathbf{D_d}$ as the terminal point reached before the first retrospection. In jigsaw puzzle and complex counting, we treat the state of each patch or region upon its first utilization as $\mathbf{D_d}$. For embodied long-horizon planning, we define $\mathbf{D_d}$ by identifying reflective adjustments in the execution path, specifically capturing repetitive actions such as re-navigating to the same location or re-picking up the same object.

\subsection{Incremental Prompting Information Injection Strategy}
Based on above three-stage evaluations and metrics, we introduce the Incremental Prompting Information Injection (IPII) strategy, formally represented as a set of hierarchical prompting levels $\mathcal{H} = \{\mathcal{P}_{\text{L1}}, \mathcal{P}_{\text{L2}}, \mathcal{P}_{\text{L3}}\}$, to ablatively explore the prompting factors for VI-CoT performance. Concretely, Level-1 establishes the baseline only utilizing the original instruction set $\mathcal{I}_{base}$, defined as $\mathcal{P}_{\text{L1}} = \mathcal{I}_{base}$. Level-2 involves implicit VI-CoT prompts by injecting a guidance term $\mathcal{I}_{imp}$, formulated as $\mathcal{P}_{\text{L2}} = \mathcal{P}_{\text{L1}} \oplus \mathcal{I}_{imp}$. This is deployed to guide the models to update its internal step-wise IVS, thereby compelling them to leverage visual imagination for subsequent planning trajectory rather than solely relying on the initial visual observation. Level-3 further augments the prompts with external knowledge $\mathcal{K}_{ext}$ to enhance visual perception, which can be represented as $\mathcal{P}_{\text{L3}} = \mathcal{P}_{\text{L2}} \oplus \mathcal{K}_{ext}$.

\section{Experiments}
\label{sec:exprs}

\begin{table*}[t!]
\centering
\caption{The performace evaluations of advanced MLLMs in terms of targeted metrics~(\%) on maze navigation. Note that we highlight the best performance in \textbf{\textit{bold}} and \underline{underline} the second performance.}
\vspace{-0.4em}
\setlength{\arrayrulewidth}{0.40pt}
\renewcommand\arraystretch{1.0}
\resizebox{\textwidth}{!}{
\begin{tabular}{lccccccccccccccc}
\toprule[0.40pt]
\multirow{2}{*}{Method} & \multirow{2}{*}{Organization} & \multirow{2}{*}{\raisebox{-2ex}{\begin{tabular}[c]{@{}c@{}}Reason- \\oriented\end{tabular}}}    
 & \multicolumn{1}{c}{Stage-1}  & \multicolumn{3}{c}{Stage-2} & \multicolumn{4}{c}{Stage-3} \\ 
\cmidrule(lr){4-4} \cmidrule(lr){5-7} \cmidrule(lr){8-11}
& &  & ACC  & ACC  & Recall\_o  & Legality  & ACC  & Recall\_o & ThinkGain & Legality   \\ 
\midrule

\multicolumn{5}{l}{\textit{\textcolor{cvprblue!80}{\textbf{Commercial proprietary MLLMs}}}} \\
GPT-4o-20240513~\cite{openai2024gpt4o}  & OpenAI  &$\times$     & 50.00 & 0.00 & 11.65 & 53.18 & 66.40 & 69.41 & \underline{33.01} & 63.76  \\
o3~\cite{openai2025o3mini} & OpenAI    & \checkmark     & 87.60 & \textbf{17.52} & \textbf{29.11} & \textbf{70.55} & \textbf{74.40} & \textbf{76.86} & \textbf{44.05} & \underline{81.33}   \\
o4-mini~\cite{openai2025o4mini} & OpenAI    & \checkmark     & \textbf{94.40} & \underline{15.45} & \underline{22.33} & \underline{68.45} & 58.40 & 61.78 & 27.98 & 69.88   \\
Gemini-2.5-pro-preview-03-25~\cite{Gemini-2.5-pro-preview-03-25} & Google   & \checkmark     & \underline{94.00} & 6.80 & 12.95 & 55.05 & \underline{68.80} & \underline{70.27} & 32.96 & 73.28   \\
Gemini-2.0-flash~\cite{Gemini-2.0-flash} & Google  & \checkmark    & 60.80 & 0.00 & 11.35 & 54.94 & 53.20 & 59.48 & 28.11 & 64.79   \\
Claude-3.5-Sonnet-20241022~\cite{anthropic2024claude3d5} & Anthropic   & \checkmark    & 53.20 & 1.60 & 11.80 & 52.07 & 25.60 & 27.90 & 6.23 & \textbf{84.63}   \\
Step-1o-vision-32k~\cite{step-1o-vision-32k} & StepFun   & $\times$     & 46.40 & 0.80 & 10.72 & 51.08 & 16.40 & 21.64 & 4.84 & 30.87   \\
Doubao-1.5-vision-pro-32k~\cite{Doubao-1.5-vision-pro-32k} & ByteDance   & $\times$     & 44.00 & 1.20 & 10.89 & 52.03 & 13.20 & 28.53 & 6.22 & 58.40   \\
SenseChat-Vision-V5.5~\cite{SenseChat-Vision-V5.5}  & SenseTime   & $\times$     & 62.80 & 0.00 & 11.22 & 53.03 & 22.40 & 34.36 & 10.50 & 63.92   \\
Hunyuan-Vision-20250103~\cite{Hunyuan-Vision-20250103} & Tencent  & $\times$  & 30.00 & 0.40 & 8.89 & 48.73 & 16.40 & 27.86 & 7.03 & 46.57   \\
\midrule

\multicolumn{5}{l}{\textit{\textcolor{cvprblue!80}{\textbf{Open-source MLLMs}}}} \\
LLaVA-OneVision-72B~\cite{Llava-onevision} & ByteDance \& NTU & $\times$    & 38.00 & 0.00 & 8.77 & 52.25 & 18.40 & 30.44 & 8.17 & 49.80  \\
InternVL-2.5-78B~\cite{Internvl} & Shanghai AI Lab & $\times$ & 46.80 & 0.00 & 10.75 & 53.47 & 29.60 & 36.31 & 11.27 & 60.52  \\
InternVL-3-78B~\cite{InternVL3} & Shanghai AI Lab & $\times$ & 43.60 & 0.00 & 11.74 & 52.73 & 37.60 & 45.03 & 12.21 & 61.86  \\
Qwen2-VL-72B-Instruct~\cite{Qwen2-vl} & Alibaba & $\times$ & 47.20 & 0.00 & 13.67 & 52.94 & 45.20 & 50.33 & 8.11 & 53.58  \\
Qwen2.5-VL-72B-Instruct~\cite{Qwen2d5-vl} & Alibaba & $\times$  & 55.60 & 1.20 & 11.27 & 49.13 & 41.20 & 45.30 & 11.77 & 57.04  \\
Qwen-Omni-Turbo~\cite{Qwen-Omni-Turbo} & Alibaba & $\times$    & 28.00 & 0.00 & 7.97 & 30.89 & 14.00 & 22.84 & 5.52 & 43.44  \\
GLM-4V-Plus~\cite{glm2024chatglm} & Zhipu AI & $\times$    & 54.40 & 0.00 & 10.53 & 51.57 & 32.80 & 37.46 & 8.08 & 50.38  \\
SkyWork-R1V1-38B~\cite{Skywork-r1v} & Skywork & \checkmark    & 36.40 & 0.40 & 7.91 & 38.49 & 3.60 & 12.55 & 0.27 & 29.34  \\
\hline
\end{tabular}
}
\vspace{-0.5em}
\label{tab:main-eval_maze}
\end{table*}

\subsection{Experimental Setup}
\label{exprs:setup}

\noindent \textbf{Evaluation Models.}~We globally select a total of 18 top-performing MLLMs for comprehensive evaluation, comprising 10 commercial proprietary models and 8 powerful open-source models. Regarding proprietary models, we include the leading OpenAI and Gemini series, specifically GPT-4o-20240513~\cite{openai2024gpt4o}, o3~\cite{openai2025o3mini}, o4-mini~\cite{openai2025o4mini}, Gemini-2.5-pro~\cite{Gemini-2.5-pro-preview-03-25}, and Gemini-2.0-flash~\cite{Gemini-2.0-flash}. We also assess popular models from other competitive organizations, including Claude-3.5-Sonnet~\cite{anthropic2024claude3d5}, Step-1o-vision-32k~\cite{step-1o-vision-32k}, Doubao-1.5-vision-pro-32k~\cite{Doubao-1.5-vision-pro-32k}, SenseChat-Vision-V5.5~\cite{SenseChat-Vision-V5.5}, as well as Hunyuan-Vision-20250103~\cite{Hunyuan-Vision-20250103}. As for open-source models, we select representative series with their largest parameter capacities to investigate performance gaps. These include LLaVA-OneVision-72B~\cite{Llava-onevision}, InternVL-2.5-78B~\cite{Internvl}, InternVL-3-78B~\cite{InternVL3}, Qwen2-VL-72B-Instruct~\cite{Qwen2-vl}, Qwen2.5-VL-72B-Instruct~\cite{Qwen2d5-vl}, Qwen-Omni-Turbo~\cite{Qwen-Omni-Turbo}, and GLM-4V-Plus~\cite{glm2024chatglm}. Finally, we evaluate MLLMs with targeted thinking capabilities, represented by SkyWork-R1V1-38B~\cite{Skywork-r1v}. Note that we take GPT-4o~\cite{openai2024gpt4o} as the baseline model for our experiments.

\noindent \textbf{Implementation Details.}~We access proprietary and open models via APIs and local deployments. The maximum token limit is 8192, temperature is 0, both Top-K and Top-P are 1. The rest of hyper-parameter settings keep same with the default settings of VLMEvalKit~\cite{Vlmevalkit}. All the experiments are conducted on a machine with 8 $\times$ NVIDIA A100 GPUs (80G). Due to space constraint, we only apply IPII strategy for maze navigation here. Specifically, Level-1 prompts are consistent with origin prompts, which are illustrated in Fig.~\textcolor{blue}{1} of supplementary material. Level-2 builds upon Level-1 by incorporating the following prompting information: \textit{Please make sure that after executing the move at each step, you should update your internal intermediate visual state, rather than remaining in the initial input visual state}, as displayed in Fig.~\textcolor{blue}{5} of supplementary material. As shown in Fig.~\textcolor{blue}{6} of supplementary material, Level-3 further builds upon Level-2 by explicitly incorporating the coordinates information of the starting and end points in maze navigation.

\begin{table*}[t!]
\centering
\caption{The performace evaluations of advanced MLLMs on jigsaw puzzle.}
\vspace{-0.4em}
\setlength{\arrayrulewidth}{0.40pt}
\renewcommand\arraystretch{1.0}
\resizebox{\textwidth}{!}{
\begin{tabular}{lccccccccccccccc}
\toprule[0.40pt]
\multirow{2}{*}{Method} & \multirow{2}{*}{Organization} & \multirow{2}{*}{\raisebox{-2ex}{\begin{tabular}[c]{@{}c@{}}Reason- \\oriented\end{tabular}}}    
 & \multicolumn{1}{c}{Stage-1}  & \multicolumn{3}{c}{Stage-2} & \multicolumn{4}{c}{Stage-3} \\ 
\cmidrule(lr){4-4} \cmidrule(lr){5-7} \cmidrule(lr){8-11}
& &  & ACC  & ACC  & Recall\_o  & Legality  & ACC  & Recall\_o & ThinkGain & Legality   \\ 
\midrule
\multicolumn{5}{l}{\textit{\textcolor{cvprblue!80}{\textbf{Commercial proprietary MLLMs}}}} \\
GPT-4o-20240513~\cite{openai2024gpt4o}  & OpenAI  &$\times$     & 26.40 & 2.00 & 28.60 & \textbf{100} & 2.00 & 31.42 & \underline{0.85} & 86.90  \\
o3~\cite{openai2025o3mini} & OpenAI    & \checkmark     & \underline{34.80} & \underline{18.00} & \underline{53.27} & 98.80 & \underline{24.00} & \underline{58.22} & \textbf{1.38} & 93.93   \\
o4-mini~\cite{openai2025o4mini} & OpenAI    & \checkmark     & 32.40 & 0.80 & 16.00 & 88.40 & 1.60 & 12.49 & -0.32 & 35.98   \\
Gemini-2.5-pro-preview-03-25~\cite{Gemini-2.5-pro-preview-03-25} & Google   & \checkmark     & \textbf{38.40} & \textbf{20.80} & \textbf{57.00} & \textbf{100} & \textbf{28.40} & \textbf{64.09} & 0.13 & 97.17   \\
Gemini-2.0-flash~\cite{Gemini-2.0-flash} & Google  & \checkmark    & 11.60 & 5.60 & 41.67 & \underline{99.93} & 10.80 & 39.17 & -0.53 & 72.13   \\
Claude-3.5-Sonnet-20241022~\cite{anthropic2024claude3d5} & Anthropic   & \checkmark    & 34.40 & 0.80 & 26.60 & 99.87 & 0.80 & 9.34 & -1.55 & 23.89   \\
Step-1o-vision-32k~\cite{step-1o-vision-32k} & StepFun   & $\times$     & 32.20 & 0.00 & 20.87 & 98.87 & 1.20 & 22.93 & -0.01 & 92.70   \\
Doubao-1.5-vision-pro-32k~\cite{Doubao-1.5-vision-pro-32k} & ByteDance   & $\times$     & 22.80 & 0.00 & 16.73 & \textbf{100} & 0.00 & 16.10 & 0.07 & 88.97   \\
SenseChat-Vision-V5.5~\cite{SenseChat-Vision-V5.5}  & SenseTime   & $\times$     & 26.40 & 1.60 & 32.87 & 98.80 & 1.60 & 31.91 & -0.26 & 97.26   \\
Hunyuan-Vision-20250103~\cite{Hunyuan-Vision-20250103} & Tencent  & $\times$ & 24.80 & 0.00 & 16.80 & 98.80 & 0.00 & 12.10 & -1.30 & 69.63   \\

\midrule

\multicolumn{5}{l}{\textit{\textcolor{cvprblue!80}{\textbf{Open-source MLLMs}}}} \\
LLaVA-OneVision-72B~\cite{Llava-onevision} & ByteDance \& NTU & $\times$    & 14.40 & 0.00 & 17.27 & \textbf{100} & 0.00 & 15.80 & 0.00 & \textbf{100}  \\
InternVL-2.5-78B~\cite{Internvl} & Shanghai AI Lab & $\times$ & 27.20 & 2.00 & 31.73 & \textbf{100} & 2.40 & 28.25 & -0.27 & 89.24  \\
InternVL-3-78B~\cite{InternVL3} & Shanghai AI Lab & $\times$ & 24.80 & 0.80 & 26.60 & \textbf{100} & 4.40 & 33.24 & -0.20 & 84.85  \\
Qwen2-VL-72B-Instruct~\cite{Qwen2-vl} & Alibaba & $\times$ & 25.20 & 0.40 & 19.47 & \textbf{100} & 0.40 & 19.48 & 0.06 & \underline{99.61}  \\
Qwen2.5-VL-72B-Instruct~\cite{Qwen2d5-vl} & Alibaba & $\times$  & 34.00 & 0.40 & 25.13 & \underline{99.93} & 0.80 & 25.75 & -0.33 & 94.25  \\
Qwen-Omni-Turbo~\cite{Qwen-Omni-Turbo} & Alibaba & $\times$    & 20.80 & 0.00 & 17.33 & 99.60 & 0.00 & 16.88 & -0.14 & 90.62  \\
GLM-4V-Plus~\cite{glm2024chatglm} & Zhipu AI & $\times$    & \underline{34.80} & 0.40 & 22.07 & \textbf{100} & 0.80 & 14.36 & 0.00 & 61.74  \\
SkyWork-R1V1-38B~\cite{Skywork-r1v} & Skywork & \checkmark    & 20.40 & 0.00 & 15.13 & 80.80 & 0.00 & 14.60 & 0.00 & 74.89  \\
\hline
\end{tabular}
}
\label{tab:main-eval_puzzle}
\end{table*}

\begin{table*}[t!]
\centering
\caption{The performace evaluations of advanced MLLMs on embodied long-horizon planning.}
\vspace{-0.4em}
\setlength{\arrayrulewidth}{0.40pt}
\renewcommand\arraystretch{1.0}
\resizebox{\textwidth}{!}{
\begin{tabular}{lccccccccccccccc}
\toprule[0.40pt]
\multirow{2}{*}{Method} & \multirow{2}{*}{Organization} & \multirow{2}{*}{\raisebox{-2ex}{\begin{tabular}[c]{@{}c@{}}Reason- \\oriented\end{tabular}}}    
 & \multicolumn{1}{c}{Stage-1}  & \multicolumn{3}{c}{Stage-2} & \multicolumn{4}{c}{Stage-3} \\ 
\cmidrule(lr){4-4} \cmidrule(lr){5-7} \cmidrule(lr){8-11}
& &  & ACC  & ACC  & Recall\_o  & Legality  & ACC  & Recall\_o & ThinkGain & Legality   \\ 
\midrule
\multicolumn{5}{l}{\textit{\textcolor{cvprblue!80}{\textbf{Commercial proprietary MLLMs}}}} \\
GPT-4o-20240513~\cite{openai2024gpt4o}  & OpenAI  &$\times$     & 57.20 & 0.04 & 0.40 & 71.13 & 12.80 & 35.07 & 22.95 & 43.74  \\
o3~\cite{openai2025o3mini} & OpenAI    & \checkmark     & 63.60 & 1.20 & 1.58 & 70.89 & \underline{16.00} & 29.48 & 20.69 & 56.57   \\
o4-mini~\cite{openai2025o4mini} & OpenAI    & \checkmark     & \textbf{68.80} & 2.40 & 3.07 & 70.77 & \textbf{22.80} & \underline{41.61} & \underline{26.14} & 53.76   \\
Gemini-2.5-pro-preview-03-25~\cite{Gemini-2.5-pro-preview-03-25} & Google   & \checkmark     & \underline{67.60} & \textbf{6.80} & 16.69 & 65.00 & \textbf{22.80} & 34.84 & 19.43 & 53.84   \\
Gemini-2.0-flash~\cite{Gemini-2.0-flash} & Google  & \checkmark    & 41.20 & 1.20 & 3.79 & 59.37 & 7.60 & 18.74 & 14.97 & 47.69   \\
Claude-3.5-Sonnet-20241022~\cite{anthropic2024claude3d5} & Anthropic   & \checkmark    & 49.20 & 0.00 & 0.11  & 42.88 & 15.00 & \textbf{43.75} & \textbf{27.08} & 59.51   \\
Step-1o-vision-32k~\cite{step-1o-vision-32k} & StepFun   & $\times$     & 61.20 & \underline{3.20} & 3.44 & 68.86 & 6.80 & 23.79 & 10.77 & 42.68   \\
Doubao-1.5-vision-pro-32k~\cite{Doubao-1.5-vision-pro-32k} & ByteDance   & $\times$     & 64.00 & 0.00 & \textbf{50.37}  & 70.97 & 14.00 & 31.17 & 19.91 & 49.54    \\
SenseChat-Vision-V5.5~\cite{SenseChat-Vision-V5.5}  & SenseTime   & $\times$     & 25.60 & 0.00 & 0.40 & 64.91 & 0.00 & 0.18 & 0.08 & 0.26   \\
Hunyuan-Vision-20250103~\cite{Hunyuan-Vision-20250103} & Tencent  & $\times$ & 22.80 & 0.00 & 0.24 & 36.76 & 3.60 & 21.49 & 14.47 & 35.40   \\

\midrule

\multicolumn{5}{l}{\textit{\textcolor{cvprblue!80}{\textbf{Open-source MLLMs}}}} \\

LLaVA-OneVision-72B~\cite{Llava-onevision} & ByteDance \& NTU & $\times$    & 48.80 & 0.00 & \underline{17.31}  & \underline{72.71} & 0.00 & 4.17 & 2.50 & 11.55  \\
InternVL-2.5-78B~\cite{Internvl} & Shanghai AI Lab & $\times$ & 39.20 & 0.00 & 0.76 & 53.83 & 0.00 & 3.69 & 2.94 & \textbf{94.72}  \\
InternVL-3-78B~\cite{InternVL3} & Shanghai AI Lab & $\times$ & 47.20 & 0.80 & 1.58 & 43.64 & 0.80 & 10.40 & 7.50 & \underline{82.98}  \\
Qwen2-VL-72B-Instruct~\cite{Qwen2-vl} & Alibaba & $\times$ & 46.40 & 1.20 & 1.76 & \textbf{74.17} & 10.80 & 25.15 & 11.45 & 28.32  \\
Qwen2.5-VL-72B-Instruct~\cite{Qwen2d5-vl} & Alibaba & $\times$  & 52.00 & 0.80 & 1.12 & 69.74 & 10.00 & 23.75 & 9.64 & 35.84  \\
Qwen-Omni-Turbo~\cite{Qwen-Omni-Turbo} & Alibaba & $\times$   & 26.40 & 0.90 & 0.00  & 0.00 & 0.71 & 6.00 & 3.51 & 7.71 \\
GLM-4V-Plus~\cite{glm2024chatglm} & Zhipu AI & $\times$    & 46.80 & 2.40 & 4.88 & 54.50 & 4.80 & 25.51 & 13.37 & 29.58  \\
SkyWork-R1V1-38B~\cite{Skywork-r1v} & Skywork & \checkmark    & 23.20 & 0.40 & 0.76 & 7.52 & 0.00 & 0.27 & 0.13 & 0.34  \\
\hline
\end{tabular}
}
\vspace{-1.0em}
\label{tab:main-eval_embodied}
\end{table*}

\begin{table*}[t!]
\centering
\caption{The performace evaluations of advanced MLLMs on complex counting.}
\vspace{-0.4em}
\setlength{\arrayrulewidth}{0.40pt}
\renewcommand\arraystretch{1.0}
\resizebox{\textwidth}{!}{
\begin{tabular}{lcccccccccccccc}
\toprule[0.40pt]
\multirow{2}{*}{Method} & \multirow{2}{*}{Organization} & \multirow{2}{*}{\raisebox{-2ex}{\begin{tabular}[c]{@{}c@{}}Reason- \\oriented\end{tabular}}}    
 & \multicolumn{1}{c}{Stage-1}  & \multicolumn{2}{c}{Stage-2} & \multicolumn{3}{c}{Stage-3} \\ 
\cmidrule(lr){4-4} \cmidrule(lr){5-6} \cmidrule(lr){7-9}
& &  & ACC  & ACC  & Recall\_o    & ACC  & Recall\_o & ThinkGain   \\ 
\midrule
\multicolumn{5}{l}{\textit{\textcolor{cvprblue!80}{\textbf{Commercial proprietary MLLMs}}}} \\
GPT-4o-20240513~\cite{openai2024gpt4o}  & OpenAI  &$\times$     & 28.80 & 12.00 & \underline{44.30}  & 14.40 & 41.50 & \underline{7.60}   \\
o3~\cite{openai2025o3mini} & OpenAI    & \checkmark     & 38.40 & 10.40 & 40.00  & \underline{17.60} & \textbf{50.70} & \textbf{10.10}   \\
o4-mini~\cite{openai2025o4mini} & OpenAI    & \checkmark     & 40.00 & 2.40 & 25.40  & 3.20 & 24.70 & 6.20   \\
Gemini-2.5-pro-preview-03-25~\cite{Gemini-2.5-pro-preview-03-25} & Google   & \checkmark     & 41.20 & \textbf{18.80} & 36.60  & \textbf{18.00} & 43.60 & 0.00   \\
Gemini-2.0-flash~\cite{Gemini-2.0-flash} & Google  & \checkmark    & \underline{46.80} & 5.60 & 22.00  & 12.80 & 33.50 & 1.40   \\
Claude-3.5-Sonnet-20241022~\cite{anthropic2024claude3d5} & Anthropic   & \checkmark    & 21.60 & 3.20 & 22.00  & 4.40 & 28.90 & 6.80   \\
Step-1o-vision-32k~\cite{step-1o-vision-32k} & StepFun   & $\times$     & 43.20 & 9.60 & 28.50  & 6.80 & 27.80 & 0.00   \\
Doubao-1.5-vision-pro-32k~\cite{Doubao-1.5-vision-pro-32k} & ByteDance   & $\times$     & 43.60 & 13.20 & 33.20  & 6.00 & 21.80 & -11.00   \\
SenseChat-Vision-V5.5~\cite{SenseChat-Vision-V5.5}  & SenseTime   & $\times$     & 24.00 & 14.80 & \textbf{44.70}  & 15.20 & \underline{44.10} & 5.10   \\
Hunyuan-Vision-20250103~\cite{Hunyuan-Vision-20250103} & Tencent  & $\times$  & 32.00 & 8.00 & 30.90  & 3.20 & 16.00 & -0.30   \\
\midrule
\multicolumn{5}{l}{\textit{\textcolor{cvprblue!80}{\textbf{Open-source MLLMs}}}} \\
LLaVA-OneVision-72B~\cite{Llava-onevision} & ByteDance \& NTU & $\times$    & 35.20 & 6.00 & 22.60  & 8.00 & 24.20 & -1.70  \\
InternVL-2.5-78B~\cite{Internvl} & Shanghai AI Lab & $\times$ & 21.20 & 6.40 & 23.00  & 4.80 & 22.70 & 0.30  \\
InternVL-3-78B~\cite{InternVL3} & Shanghai AI Lab & $\times$ & 28.80 & 11.60 & 34.90  & 15.20 & 37.00 & 0.00  \\
Qwen2-VL-72B-Instruct~\cite{Qwen2-vl} & Alibaba & $\times$  & 36.00 & 11.60 & 36.20  & 12.40 & 33.00 & -0.60  \\
Qwen2.5-VL-72B-Instruct~\cite{Qwen2d5-vl} & Alibaba & $\times$  & \textbf{50.00} & 8.00 & 32.90  & 10.00 & 29.70 & 0.00  \\
Qwen-Omni-Turbo~\cite{Qwen-Omni-Turbo} & Alibaba & $\times$   & 27.60 & \underline{16.40} & 44.00  & 10.00 & 26.60 & -7.90  \\
GLM-4V-Plus~\cite{glm2024chatglm} & Zhipu AI & $\times$    & 37.20 & 14.40 & 43.00  & 9.20 & 37.10 & 0.00  \\
SkyWork-R1V1-38B~\cite{Skywork-r1v} & Skywork & \checkmark    & 36.80 & 6.40 & 27.30  & 7.60 & 26.70 & 0.00  \\
\hline
\end{tabular}
}
\vspace{-0.3em}
\label{tab:main-eval_counting}
\end{table*}

\subsection{Main Results}

\begin{figure*}[t!]
\centering
\includegraphics[width=\linewidth]{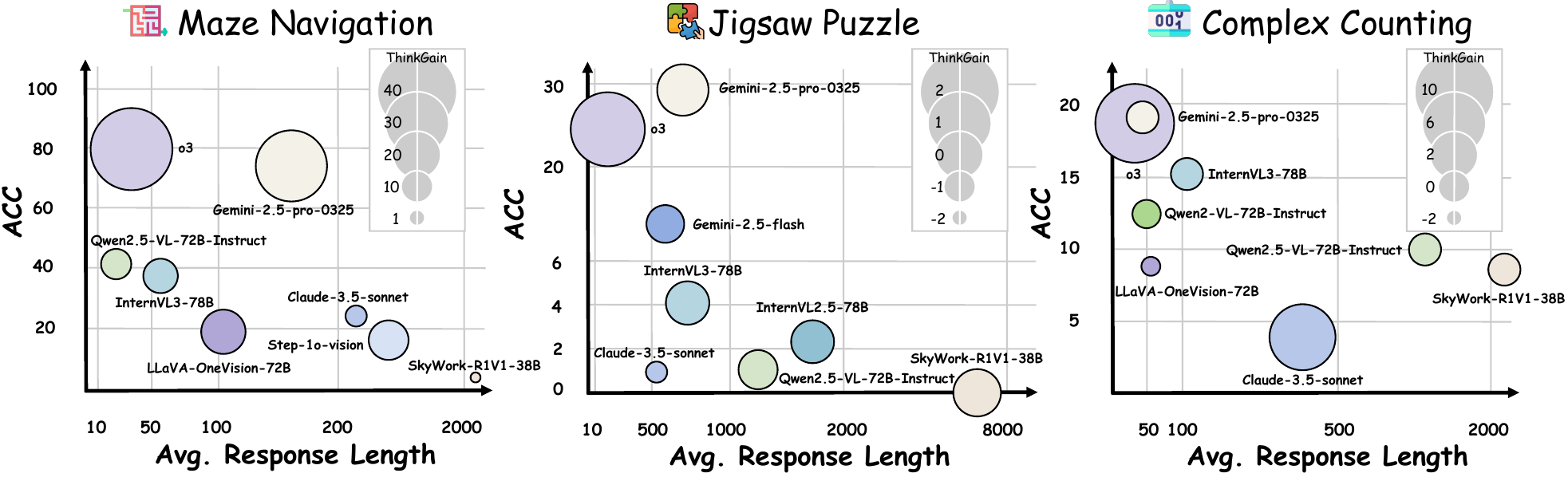}
\vspace{-1.4em}
\caption{The visualized comparisons on averge response length, ACC, and ThinkGain for three tasks.}
\vspace{-1.2em}
\label{fig:gain}
\end{figure*}

\noindent \textbf{Maze Navigation.}~Tab.~\ref{tab:main-eval_maze} indicates that most MLLMs exhibit competent performance in Stage 1. Performance significantly drops in Stage 2, indicating that current MLLMs have limitations in open-ended spatial reasoning and perception. In Stage 3, with the supports of free-style VIS, all models consistently achieves gains in global-level ACC and fine-grained $\mathbf{R\_o}$, leading to impressive ThinkGain, which indicates the effectiveness of free-style IVS in tackling deficiencies of spatial-aware cognition. However, we observe decline in Legality, which indicates that external knowledge could further confuse the thinking trajectories of weaker models. From Fig.~\ref{fig:gain}, we can clearly observe that response length is inversely proportional to efficiency in maze navigation, implying that excessive verbosity often signals uncertainty or error accumulation. These outcomes suggest that free-style IVS can act as a critical visual anchor, allowing MLLMs with strong priors to verify intermediate states and significantly enhance the spatial-aware reasoning performance.

\noindent \textbf{Jigsaw Puzzle.}~Tab.~\ref{tab:main-eval_puzzle} indicates that there also exists significant declines from Stage 1 to 2, confirming that open-ended QA poses challenges to visual CoT in this task, which is likely to stem from the difficulties in understanding AIGC semantics and distinguishing semantically incoherent patches. MLLMs in Stage 3 promoted by free-style IVS exhibit global-level gains compared to Stage 2, but surprisingly achieve negative ThinkGain with impressive drops in Legality, which could be due to the irrational IVS utilizations leading to invalid decisions. Fig.~\ref{fig:gain} indicates that top-performing models~\cite{openai2025o3mini,Gemini-2.5-pro-preview-03-25} achieve high efficiency with concise responses. In stark contrast, SkyWork-R1V1~\cite{Skywork-r1v} suffers from performance wipeout despite its lengthy reasoning chains, indicating that verbose CoT does not guarantee success without effective grounding. These failures underscore the deficiencies of current MLLMs in effectively integrating free-form IVS, revealing an immature VI-CoT capability where visual feedback may distract rather than guide. Consequently, future works should explore training paradigms that can effectively align visual generation with reasoning goals to ensure the positive utilization of IVS.

\noindent \textbf{Embodied Long-Horizon Planning.}~Most models exhibit significant declines from Stage 1 to Stage 2, exemplified by GPT-4o plummeting from $57.2\%$ to $0.04\%$. This exposes that current MLLMs possess a sophisticated linguistic shell but lack grounded physical world capability, leading to embodied hallucinations that defy basic laws. Deprived of options, models degenerate into blind guessing, with Skywork-R1V1~\cite{Skywork-r1v} even failing basic instruction constraints with $7.52\%$ Legality metric. Crucially, this performance collapse is generally reversed in Stage 3, particularly for advanced models. The substantial ThinkGain metric demonstrates that free-style IVS serves as a vital visual anchor, empowering capable models to transform blind hallucinations into verifiable actions, thereby reactivating the embodied reasoning potential dormant in text-only contexts. For Stage 3, we further conduct extensive qualitative analysis in a real-world scenario, as illustrated in Fig.~\ref{fig:embodied-case1} below. This task poses a significant challenge as the targeted object is occluded, requiring models to infer implicit sub-goals for object searching. As observed, proprietary MLLMs~\cite{openai2024gpt4o,Gemini-2.5-pro-preview-03-25} demonstrate robust reasoning-for-planning capabilities. They successfully decompose the high-level instruction into actionable steps, and exhibit self-correction behaviors when initial attempts fail. In contrast, open-source model~\cite{Qwen2d5-vl} struggles with long-horizon dependencies, which exhibits failure modes including: (1) invalid navigation planning, where the model attempts to navigate directly to an invisible target; (2) task deviation, such as interacting with irrelevant objects; and (3) recursive behavior, getting trapped in repetitive loops of opening and closing the microwave. These comparisons underscore the critical role of free-style IVS in bridging the gap between high-level instructions and embodied execution, enabling models to maintain logical consistency over long horizons, while also revealing the significant performance gaps between open-source and closed-source models.

\noindent \textbf{Complex Counting.}~ACC in Stage 2 is inferior to Stage 1, which confirms our conclusions in the above tasks and reinforces that the removal of option-based hints exposes the inability of models to perform autonomous enumeration. In Stage 3, some models~\cite{Doubao-1.5-vision-pro-32k,Qwen-Omni-Turbo,Hunyuan-Vision-20250103} exhibit ACC declines and negative ThinkGain. Surprisingly, Gemini-2.5-pro~\cite{Gemini-2.5-pro-preview-03-25} also exhibits drops with null ThinkGain. These outcomes indicate that MLLMs still suffer from object hallucinations and deficiencies in basic perception, where low-quality generated visuals act as noise rather than valid references, leading to accumulated erroneous trajectories. In case studies, we also find that IVS assist models in recognizing heads, yet they still make wrong decisions, consistent with observations in LLMs~\cite{counting1,counting2}. As can be seen from Fig.~\ref{fig:gain}, we observe that models with relatively better performance tend to produce shorter, more concise responses. This trend suggests that capable models can efficiently identify and count objects without resorting to verbose, redundant reasoning often indicative of uncertainty. These outcomes highlight the significant challenges MLLMs still face with complex counting, particularly in maintaining spatial consistency during long-horizon enumeration. As a result, we claim that future research should focus on developing targeted representations, such as explicit visual markers, or specialized training strategies to enhance object discrimination and spatial awareness.

\begin{figure*}[t!]
\centering
\includegraphics[width=\linewidth]{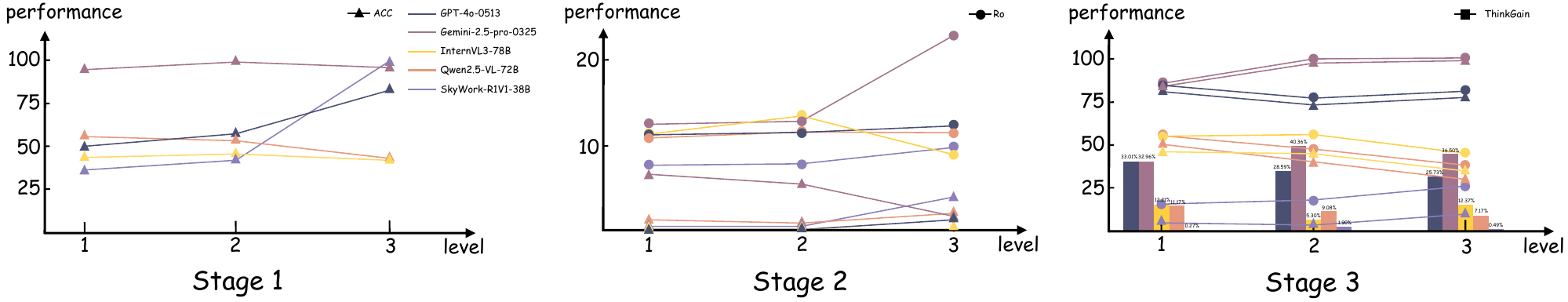}
\vspace{-1.4em}
\caption{The further investigations with IPII strategy in VI-CoT capability of advanced MLLMs for maze navigation.}
\vspace{-1.1em}
\label{fig:IPII}
\end{figure*}

\subsection{Further Investigations with IPII Strategy.}
\label{sec:further_investigation}

Our analysis across the three stages of IPII strategy demonstrates the distinct capabilities and limitations of MLLMs. In Stage 1, proprietary models consistently exhibit superior performance stability, while open-source models~\cite{InternVL3,Qwen2d5-vl} show a decline with increasing difficulty. The notable exception of SkyWork-R1V1~\cite{Skywork-r1v}, which shows a sharp accuracy surge at Level 3, suggests that R1-like MLLMs might have higher sensitivities to prompts. In Stage 2, the widespread failure to achieve significant performance gains across all models highlights a critical bottleneck, \ie existing MLLMs might lack the capability to implicitly update their internal IVS through textual CoT in the open-ended QA. These results also indicate the necessity of utilizing free-style IVS as external knowledge to improve reasoning. Finally, in Stage 3, the benefit of introducing external free-style IVS leads to a clear performance divergence. Advanced MLLMs~\cite{Gemini-2.5-pro-preview-03-25} can successfully leverage the visual contexts to boost planning accuracy, whereas weaker MLLMs often struggle with information overload, leading to confusion and decreased performance.

\begin{figure*}[t!]
\centering
\includegraphics[width=\linewidth]{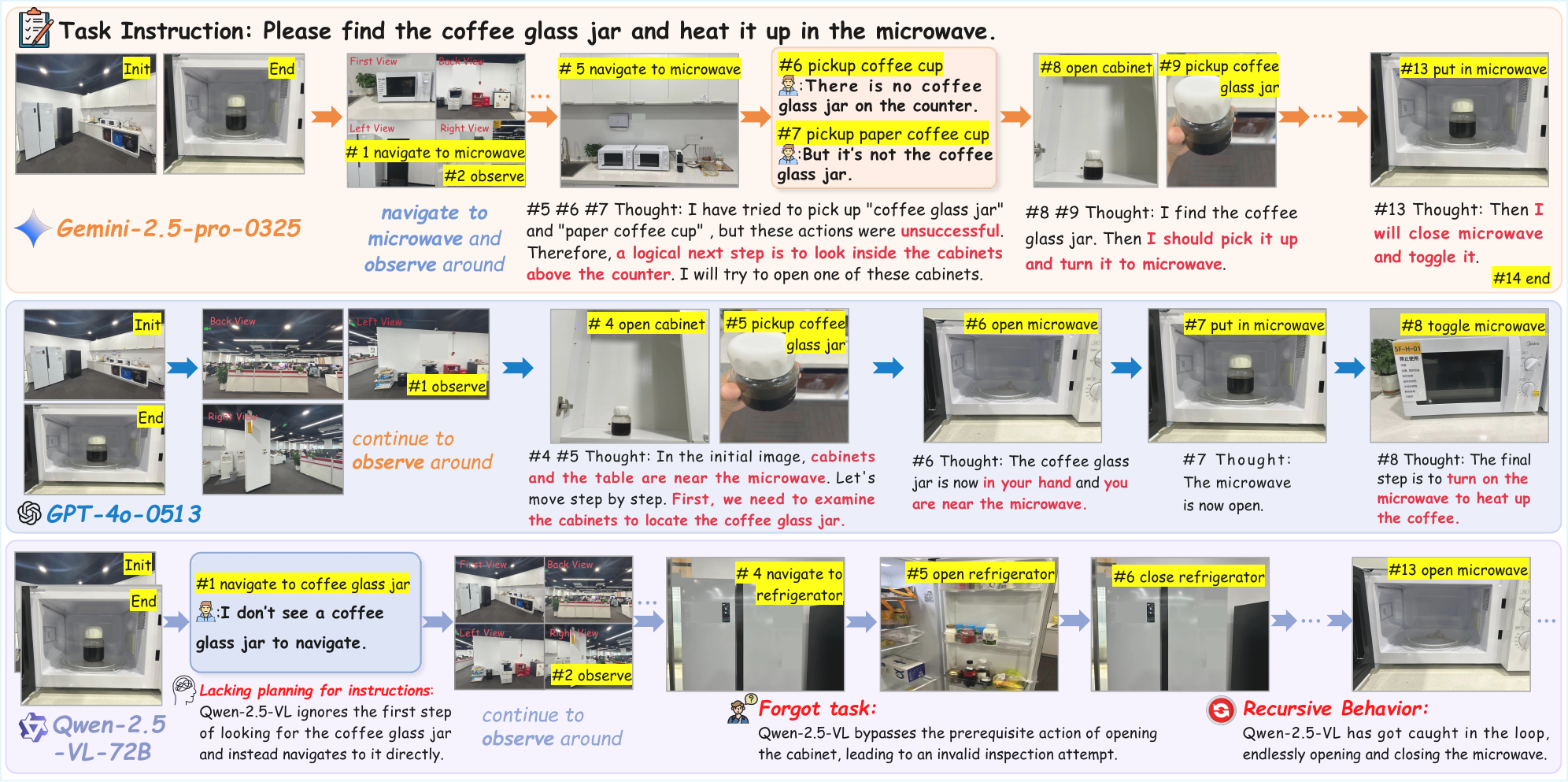}
\vspace{-1.0em}
\caption{The qualitative analysis for Stage 3 with free-style IVS in embodied long-horizon planning.}
\vspace{-1.2em}
\label{fig:embodied-case1}
\end{figure*}
\section{Conclusion and Limitations}
\label{sec:Conclusions}
We introduce ViC-Bench, a specialized benchmark designed to evaluate VI-CoT capability in MLLMs. This benchmark consists of four representative tasks, with each has dedicated construction and free-style IVS generation pipelines supporting function calls. To obtain a thorough understanding of VI-CoT performance, we design a novel progressive three-stage evaluation suite with targeted new metrics. The IPII strategy also impressively indicate the prompting factors which affect VI-CoT performance. The systematic evaluations obtain key observations and insights into the current developments of VI-CoT in MLLMs. We hope ViC-Bench can inspire more research in multi-modal interleaved reasoning. As the field continues to evolve, we hope that ViC-Bench can stand as a valuable tool for measuring progress in the development of more sophisticated multi-modal AI systems.

Despite our efforts, limitations still exist. The \textit{ThinkGain} metric involves black-boxing the retrospection of MLLMs, and we plan to deeply delve into retrospection for more detailed investigations in the future developments.

\bibliographystyle{IEEEtran}
\bibliography{main}

\end{document}